\documentclass[twocolumn]{article}
\usepackage{fullname}

\newcounter{sentencectr}
\newcounter{sentencesubctr}

\renewcommand{\thesentencectr}{(\smainform{sentencectr})}
\renewcommand{\thesentencesubctr}{\thesentencectr\ssubform{sentencesubctr}}

\newcommand{\smainform}{\arabic}
\newcommand{\ssubform}{\alph}
\newcommand{\ssubpunc}{.{}}

\newcommand{\beginsentences}{ %
\pagebreak[3] %
\begin{list}{(\thesentencectr)}
   {\usecounter{sentencesubctr}
    \setlength{\topsep}{1ex}			
    \setlength{\itemsep}{0 in}
    \setlength{\labelwidth}{0.5 in}
    \addtolength{\leftmargin}{4ex}
    \setlength{\labelsep}{.05in}
    \setlength{\parsep}{0 in}}}
\def\endsentences{\end{list}}

\newcommand{\sitem}{\renewcommand{\thesentencesubctr}{(\smainform{sentencectr}}
                    \refstepcounter{sentencectr}
     \item[(\smainform{sentencectr})\hfill]}
\newcommand{\smainitem}{\renewcommand{\thesentencesubctr
                                    }{\thesentencectr\ssubform{sentencesubctr}}
                        \setcounter{sentencesubctr}{0}
                        \refstepcounter{sentencectr}
                        \refstepcounter{sentencesubctr}
     \item[\thesentencectr\hfill\ssubform{sentencesubctr}\ssubpunc]}
\newcommand{\ssubitem}{\refstepcounter{sentencesubctr}
     \item[\hfill\ssubform{sentencesubctr}\ssubpunc]}

\makeatletter            
\newcommand{\smainlabel}[1]{{
\renewcommand{\@currentlabel}{\thesentencectr}\label{#1}}}

\newcommand{\ssublabel}[1]{{
\renewcommand{\@currentlabel}{\ssubform{sentencesubctr}}\label{#1}}}
\makeatother

\makeatother

\catcode`\@=11\relax
\newwrite\@unused
\def\typeout#1{{\let\protect\string\immediate\write\@unused{#1}}}
\def\figurepath{./}

\def\@nnil{\@nil}
\def\@empty{}
\def\@psdonoop#1\@@#2#3{}
\def\@psdo#1:=#2\do#3{\edef\@psdotmp{#2}\ifx\@psdotmp\@empty \else
    \expandafter\@psdoloop#2,\@nil,\@nil\@@#1{#3}\fi}
\def\@psdoloop#1,#2,#3\@@#4#5{\def#4{#1}\ifx #4\@nnil \else
       #5\def#4{#2}\ifx #4\@nnil \else#5\@ipsdoloop #3\@@#4{#5}\fi\fi}
\def\@ipsdoloop#1,#2\@@#3#4{\def#3{#1}\ifx #3\@nnil 
       \let\@nextwhile=\@psdonoop \else
      #4\relax\let\@nextwhile=\@ipsdoloop\fi\@nextwhile#2\@@#3{#4}}
\def\@tpsdo#1:=#2\do#3{\xdef\@psdotmp{#2}\ifx\@psdotmp\@empty \else
    \@tpsdoloop#2\@nil\@nil\@@#1{#3}\fi}
\def\@tpsdoloop#1#2\@@#3#4{\def#3{#1}\ifx #3\@nnil 
       \let\@nextwhile=\@psdonoop \else
      #4\relax\let\@nextwhile=\@tpsdoloop\fi\@nextwhile#2\@@#3{#4}}
\def\psdraft{
	\def\@psdraft{0}
}
\def\psfull{
	\def\@psdraft{100}
}
\psfull
\newif\if@prologfile
\newif\if@postlogfile
\newif\if@noisy
\def\pssilent{
	\@noisyfalse
}
\def\psnoisy{
	\@noisytrue
}
\psnoisy
\newif\if@bbllx
\newif\if@bblly
\newif\if@bburx
\newif\if@bbury
\newif\if@height
\newif\if@width
\newif\if@scale
\newif\if@rheight
\newif\if@rwidth
\newif\if@clip
\newif\if@verbose
\def\@p@@sclip#1{\@cliptrue}

\def\@p@@sfile#1{\def\@p@sfile{null}%
	        \openin1=#1
		\ifeof1\closein1%
		       \openin1=\figurepath#1
			\ifeof1\typeout{Error, File #1 not found}
			\else\closein1
			    \edef\@p@sfile{\figurepath#1}%
                        \fi%
		 \else\closein1%
		       \def\@p@sfile{#1}%
		 \fi}
\def\@p@@sfigure#1{\def\@p@sfile{null}%
	        \openin1=#1
		\ifeof1\closein1%
		       \openin1=\figurepath#1
			\ifeof1\typeout{Error, File #1 not found}
			\else\closein1
			    \def\@p@sfile{\figurepath#1}%
                        \fi%
		 \else\closein1%
		       \def\@p@sfile{#1}%
		 \fi}

\def\@p@@sbbllx#1{
		\@bbllxtrue
		\dimen100=#1
		\edef\@p@sbbllx{\number\dimen100}
}
\def\@p@@sbblly#1{
		\@bbllytrue
		\dimen100=#1
		\edef\@p@sbblly{\number\dimen100}
}
\def\@p@@sbburx#1{
		\@bburxtrue
		\dimen100=#1
		\edef\@p@sbburx{\number\dimen100}
}
\def\@p@@sbbury#1{
		\@bburytrue
		\dimen100=#1
		\edef\@p@sbbury{\number\dimen100}
}
\def\@p@@sscale#1{
		\@scaletrue
		\count255=#1
   		\edef\@p@sscale{\number\count255}
}
\def\@p@@sheight#1{
		\@heighttrue
		\dimen100=#1
   		\edef\@p@sheight{\number\dimen100}
}
\def\@p@@swidth#1{
		\@widthtrue
		\dimen100=#1
		\edef\@p@swidth{\number\dimen100}
}
\def\@p@@srheight#1{
		\@rheighttrue
		\dimen100=#1
		\edef\@p@srheight{\number\dimen100}
}
\def\@p@@srwidth#1{
		\@rwidthtrue
		\dimen100=#1
		\edef\@p@srwidth{\number\dimen100}
}
\def\@p@@ssilent#1{ 
		\@verbosefalse
}
\def\@p@@sprolog#1{\@prologfiletrue\def\@prologfileval{#1}}
\def\@p@@spostlog#1{\@postlogfiletrue\def\@postlogfileval{#1}}
\def\@cs@name#1{\csname #1\endcsname}
\def\@setparms#1=#2,{\@cs@name{@p@@s#1}{#2}}
\def\ps@init@parms{
		\@bbllxfalse \@bbllyfalse
		\@bburxfalse \@bburyfalse
		\@heightfalse \@widthfalse
		\@scalefalse
		\@rheightfalse \@rwidthfalse
		\def\@p@sbbllx{}\def\@p@sbblly{}
		\def\@p@sbburx{}\def\@p@sbbury{}
		\def\@p@sheight{}\def\@p@swidth{}
		\def\@p@sscale{}
		\def\@p@srheight{}\def\@p@srwidth{}
		\def\@p@sfile{}
		\def\@p@scost{10}
		\def\@sc{}
		\@prologfilefalse
		\@postlogfilefalse
		\@clipfalse
		\if@noisy
			\@verbosetrue
		\else
			\@verbosefalse
		\fi
}
\def\parse@ps@parms#1{
	 	\@psdo\@psfiga:=#1\do
		   {\expandafter\@setparms\@psfiga,}}
\newif\ifno@bb
\newif\ifnot@eof
\newread\ps@stream
\def\bb@missing{
	\if@verbose{
		\typeout{psfig: searching \@p@sfile \space  for bounding box}
	}\fi
	\openin\ps@stream=\@p@sfile
	\no@bbtrue
	\not@eoftrue
	\catcode`\%=12
	\loop
		\read\ps@stream to \line@in
		\global\toks200=\expandafter{\line@in}
		\ifeof\ps@stream \not@eoffalse \fi
		\@bbtest{\toks200}
		\if@bbmatch\not@eoffalse\expandafter\bb@cull\the\toks200\fi
	\ifnot@eof \repeat
	\catcode`\%=14
}	
\catcode`\%=12
\newif\if@bbmatch
\def\@bbtest#1{\expandafter\@a@\the#1
\long\def\@a@#1
\long\def\bb@cull#1 #2 #3 #4 #5 {
	\dimen100=#2 bp\edef\@p@sbbllx{\number\dimen100}
	\dimen100=#3 bp\edef\@p@sbblly{\number\dimen100}
	\dimen100=#4 bp\edef\@p@sbburx{\number\dimen100}
	\dimen100=#5 bp\edef\@p@sbbury{\number\dimen100}
	\no@bbfalse
}
\catcode`\%=14
\def\compute@bb{
		\no@bbfalse
		\if@bbllx \else \no@bbtrue \fi
		\if@bblly \else \no@bbtrue \fi
		\if@bburx \else \no@bbtrue \fi
		\if@bbury \else \no@bbtrue \fi
		\ifno@bb \bb@missing \fi
		\ifno@bb \typeout{FATAL ERROR: no bb supplied or found}
			\no-bb-error
		\fi
		\count203=\@p@sbburx
		\count204=\@p@sbbury
		\advance\count203 by -\@p@sbbllx
		\advance\count204 by -\@p@sbblly
		\edef\@bbw{\number\count203}
		\edef\@bbh{\number\count204}
}
\def\in@hundreds#1#2#3{\count240=#2 \count241=#3
		     \count100=\count240	
		     \divide\count100 by \count241
		     \count101=\count100
		     \multiply\count101 by \count241
		     \advance\count240 by -\count101
		     \multiply\count240 by 10
		     \count101=\count240	
		     \divide\count101 by \count241
		     \count102=\count101
		     \multiply\count102 by \count241
		     \advance\count240 by -\count102
		     \multiply\count240 by 10
		     \count102=\count240	
		     \divide\count102 by \count241
		     \count200=#1\count205=0
		     \count201=\count200
			\multiply\count201 by \count100
		 	\advance\count205 by \count201
		     \count201=\count200
			\divide\count201 by 10
			\multiply\count201 by \count101
			\advance\count205 by \count201
		     \count201=\count200
			\divide\count201 by 100
			\multiply\count201 by \count102
			\advance\count205 by \count201
		     \edef\@result{\number\count205}
}
\def\compute@wfromh{
		\in@hundreds{\@p@sheight}{\@bbw}{\@bbh}
		\edef\@p@swidth{\@result}
}
\def\compute@hfromw{
		\in@hundreds{\@p@swidth}{\@bbh}{\@bbw}
		\edef\@p@sheight{\@result}
}
\def\compute@wfroms{
		\in@hundreds{\@p@sscale}{\@bbw}{100}
		\edef\@p@swidth{\@result}
}
\def\compute@hfroms{
		\in@hundreds{\@p@sscale}{\@bbh}{100}
		\edef\@p@sheight{\@result}
}
\def\compute@handw{
		\if@scale
			\compute@wfroms
			\compute@hfroms
		\else
			\if@height 
				\if@width
				\else
					\compute@wfromh
				\fi	
			\else 
				\if@width
					\compute@hfromw
				\else
					\edef\@p@sheight{\@bbh}
					\edef\@p@swidth{\@bbw}
				\fi
			\fi
		\fi
}
\def\compute@resv{
		\if@rheight \else \edef\@p@srheight{\@p@sheight} \fi
		\if@rwidth \else \edef\@p@srwidth{\@p@swidth} \fi
}
\def\compute@sizes{
	\compute@bb
	\compute@handw
	\compute@resv
}
\def\psfig#1{\vbox {
	\ps@init@parms
	\parse@ps@parms{#1}
	\compute@sizes
	\ifnum\@p@scost<\@psdraft{
		\if@verbose{
			\typeout{psfig: including \@p@sfile \space }
		}\fi
		\special{ps::[begin] 	\@p@swidth \space \@p@sheight \space
				\@p@sbbllx \space \@p@sbblly \space
				\@p@sbburx \space \@p@sbbury \space
				startTexFig \space }
		\if@clip{
			\if@verbose{
				\typeout{(clip)}
			}\fi
			\special{ps:: doclip \space }
		}\fi
		\if@prologfile
		    \special{ps: plotfile \@prologfileval \space } \fi
		\special{ps: plotfile \@p@sfile \space }
		\if@postlogfile
		    \special{ps: plotfile \@postlogfileval \space } \fi
		\special{ps::[end] endTexFig \space }
		\vbox to \@p@srheight true sp{
			\hbox to \@p@srwidth true sp{
				\hss
			}
		\vss
		}
	}\else{
		\vbox to \@p@srheight true sp{
		\vss
			\hbox to \@p@srwidth true sp{
				\hss
				\if@verbose{
					\@p@sfile
				}\fi
				\hss
			}
		\vss
		}
	}\fi
}}
\def\psglobal{\typeout{psfig: PSGLOBAL is OBSOLETE; use psprint -m instead}}
\catcode`\@=12\relax

\newcommand{\sem}[1]{\mbox{\em #1$'$}}
\newcommand{\form}[1]{\mbox{\sc\sf #1}}
\newtheorem{thm}{Theorem}
\newtheorem{defn}[thm]{Definition}

\title{Separating Dependency from Constituency in a Tree Rewriting
  System\thanks{Thanks to Christy Doran, Aravind Joshi, Nobo Komagata,
    Owen Rambow, and B. Srinivas for their helpful comments and
    discussion.}}
\author{Anoop Sarkar\\
  Department of Computer and Information Science \\
  University of Pennsylvania \\
  200 South 33rd St, Philadelphia PA 19104 \\
  {\tt anoop@linc.cis.upenn.edu}} \date{}

\begin{document}
\sloppy
\maketitle

\section{Introduction}

We define a new grammar formalism called Link-Sharing Tree Adjoining
Grammar (\form{LSTAG}) which arises directly out of a concern for
distinguishing the notion of constituency from the notion of relating
lexical items in terms of linguistic dependency%
\footnote{ The term dependency is used here broadly to include formal
  relationships such as case and agreement and other relationships
  such as filler-gap.}%
\cite{melcuk88,rambowjoshi92}. This work derives directly from work on
Tree Adjoining Grammars (\form{TAG}) \cite{joshi75} where these two
notions are conflated.  The set of derived trees for a TAG correspond
to the traditional notions of constituency while the derivation trees
of a TAG are closely related to dependency structure
\cite{rambowjoshi92}.  A salient feature of \form{TAG} is the extended
domain of locality it provides for stating these dependencies.  Each
elementary tree can be associated with a lexical item giving us a
lexicalized \form{TAG} (\form{LTAG})\cite{joshi-schabes91}.
Properties related to the lexical item such as subcategorization,
agreement, and certain types of word-order variation can be expressed
directly in the elementary tree \cite{kroch87,frank92}. Thus, in an
\form{LTAG} all of these linguistic dependencies are expressed locally
in the elementary trees of the grammar. This means that the predicate
and its arguments are always topologically situated in the same
elementary tree.

However, in coordination of predicates, e.g. \ref{ex:examples}, the
dependencies between predicate and argument cannot be represented in a
\form{TAG} elementary tree directly, since several elementary trees
seem to be `sharing' their arguments.

\beginsentences
\smainitem Kiki frolics, sings and plays all day.
\ssubitem  Kiki likes and Bill thinks Janet likes soccer.
\smainlabel{ex:examples}
\endsentences

The idea behind \form{LSTAG} is that the non-local nature of
coordination as in \ref{ex:examples} (for \form{TAG}-like grammar
formalisms) can be captured by introducing a restricted degree of {\em
  synchronized} parallelism into the \form{TAG} rewriting system while
retaining the existing {\em independent} parallelism%
\footnote{It is important to note that while the adjunction operation
  in \form{TAG}s is ``context-free'', synchronized parallelism could
  be attributed to the \form{TAG} formalism due to the string wrapping
  capabilities of adjunction, since synchronized parallelism is
  concerned with how strings are derived in a rewriting system. We
  note this as a conjecture but will not attempt to prove it
  here.}%
\cite{engelfriet:etal80,sattarambow:toappear}. We believe that an
approach towards coordination that explicitly distinguishes the
dependencies from the constituency gives a better formal understanding
of its representation when compared to previous approaches that use
tree-rewriting systems which conflate the two issues, as in
\cite{joshi90,Joshi91,sarkarjoshi96} which have to represent sentences
such as \ref{ex:examples} with either unrooted trees or by performing
structure merging on the derived tree. Other formalisms for
coordination have similar motivations: however their approaches
differ, e.g. CCG~\cite{steedman85,steedman:97} extends the notion of
constituency, while generative syntacticians~\cite{moltmann92,muadz91}
work with three-dimensional syntactic trees.

\section{Synchronized Parallelism}

The terms {\em synchronized} parallelism and {\em independent}
parallelism arise from work done on a family of formalisms termed
parallel rewriting systems that extend context-free grammars
(\form{CFG}) by the addition of various restrictive devices (see
\cite{engelfriet:etal80})). Synchronized parallelism allows
derivations which include substrings which have been generated by a
common (or shared) underlying derivation process%
\footnote{ The Lindenmayer systems are examples of systems with only
  synchronous parallelism and it is interesting to note that these $L$
  systems have the anti-AFL property
(where none of the standard closures apply).}%
. Independent parallelism corresponds to the instantiations of
independent derivation processes which are then combined to give the 
entire derivation of a string%
\footnote{ \form{CFG} is a formalism that only has
  independent parallelism. }%
. What we are exploring in this paper is an example of a mixed system
with both independent and synchronous parallelism.

In \cite{sattarambow:toappear} it is shown that by allowing an
unbounded degree of synchronized parallelism we get systems that are
too unconstrained. However, interesting subfamilies arise when the
synchronous parallelism is bounded to a finite degree, i.e. only a
bounded number of subderivations can be synchronized in a given
grammar. The system we define has this property.

\section{\form{LSTAG}}

We first look at the formalism of Synchronous \form{TAG}
(\form{STAG})\cite{shieber90} since it is an example of a
tree-rewriting system that has synchronized parallelism. 

\begin{figure*}[htbp]
  \begin{center}
    \leavevmode
    \psfig{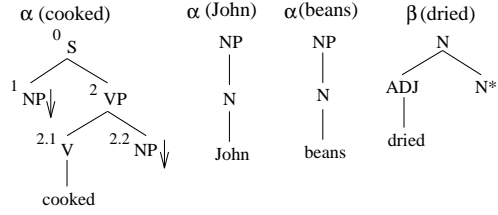}
  \end{center}
  \caption{Example of a \form{TAG}}
  \label{fig:tags}
\end{figure*}

\begin{figure*}[htbp]
  \begin{center}
    \leavevmode
    \psfig{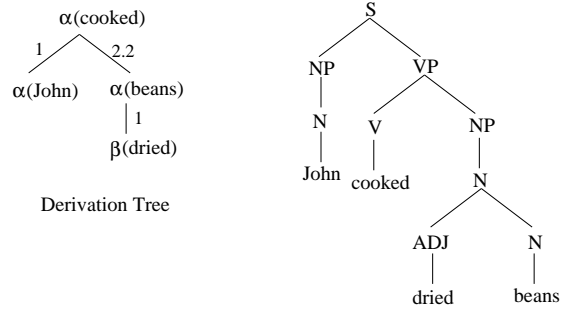}
  \end{center}
  \caption{Example of a derivation tree and corresponding parse tree}
  \label{fig:tagderiv}
\end{figure*}

As a preliminary we first informally define Tree Adjoining Grammars
(\form{TAG}).  For example, Figure~\ref{fig:tags} shows an example of
a tree for a transitive verb {\em cooked}. Each node in the tree has a
unique address obtained by applying a Gorn tree addressing scheme. For
instance, the object {\em NP} has address $2.2$. In the \form{TAG}
formalism, trees can be composed using the two operations of {\em
  substitution\/} (corresponds to string concatenation) and {\em
  adjunction\/} (corresponds to string wrapping). A history of these
operations on elementary trees in the form of a derivation tree can be
used to reconstruct the derivation of a string recognized by a
\form{TAG}.  Figure~\ref{fig:tagderiv} shows an example of a
derivation tree and the corresponding parse tree for the derived
structure obtained when $\alpha(John)$ and $\alpha(beans)$ substitute
into $\alpha(cooked)$ and $\beta(dried)$ adjoins into $\alpha(beans)$
giving us a derivation tree for {\em John cooked dried beans}. Trees
that adjoin are termed as {\em auxiliary trees}, trees that are not
auxiliary are called {\em initial}. Each node in the derivation tree
is the name of an elementary tree. The labels on the edges denote the
address in the parent node where a substitution or adjunction has
occured.

\begin{defn}
  In a \form{TAG} $G = \{ \gamma \mid \gamma$ is either an {\em
    initial} tree or an {\em auxiliary} tree $\}$, we will notate
  adjunction (similarly substitution) of trees $\gamma_1 \ldots
  \gamma_k$ into tree $\gamma$ at addresses $a_1 \ldots a_k$ giving a
  derived tree $\gamma'$ as \[ \gamma' = \gamma[a_1, \gamma_1] \ldots
  [a_k, \gamma_k]
\]
\end{defn}

\begin{defn}
  Given two standard \form{TAG}s $G_L$ and $G_R$ we define (from
  \cite{shieber94}) a \form{STAG} as $\{ \langle \gamma, \gamma',
  \frown \rangle \mid \gamma \in G_L, \gamma' \in G_R \}$, where
  $\frown$ is a set of {\em links} from a node address in $\gamma$ to
  a node address in $\gamma'$. A derivation proceeds as follows:
\begin{itemize}
\item for $\gamma = \langle \gamma_L, \gamma_R, \frown \rangle$, pick
  a link member $a_L \frown_i a_R$, where the $a$'s are node addresses
  and $\frown_i\ \in\ \frown$. For simplicity, we refer to $\frown$ as
  {\em link} and its elements $\frown_i$ as link {\em members}.
\item adjunction (similarly substitution) of $\langle \beta_L,
  \beta_R, \frown' \rangle$ into $\gamma$ is given by \[ \langle
  \gamma'_L, \gamma'_R, \frown'' \rangle = \langle \gamma_L[a_L,
  \beta_L] , \gamma_R[a_R, \beta_R], \frown'' \rangle\]
  where all links in $\frown$ and $\frown'$ are included in $\frown''$
  except $\frown_i$.
\item \( \langle \gamma'_L, \gamma'_R, \frown'' \rangle \) is now a
  derived structure which can be further operated upon.
\end{itemize}
\label{defn:STAG}
\end{defn}

In~\cite{abeille92,abeille94} \form{STAG}s have been used in handling
non-local dependencies and to seperate syntactic attachment from
semantic roles. However, \form{STAG} cannot be used to seperate the
dependencies created in (pairs of) derivation trees for coordinate
structures from the constituency represented in these derivation
trees. In this particular sense, \form{STAG} has the same shortcomings
of a \form{TAG}.  Also the above definition of the inheritance of
links in derived structures allows \form{STAG} to derive strings not
generable by \form{TAG} \cite{shieber94}. We look at a modified
version of \form{STAG}s which is weaker in power than \form{STAG}s as
defined in Defn~\ref{defn:STAG}.  We call this formalism {\em
  Link-Sharing}~\form{TAG} (\form{LSTAG}).

\begin{defn}
  An \form{LSTAG} $G$ is defined as a 4-tuple $\langle G_L, G_R,
  \Delta, \Phi \rangle$ where $G_L, G_R$ are standard \form{TAG}s,
  $\Delta$ and $\Phi$ are disjoint sets of sets of links and for each
  pair $\gamma = \langle \gamma_L, \gamma_R \rangle$, where $\gamma_L
  \in G_L$ and $\gamma_R \in G_R$, $\delta_\gamma \in \Delta$ is a
  subset of {\em links} in $\gamma$ and $\phi_{\gamma_R} \in \Phi$ is
  a distinguished subset of {\em links} with the following properties:

\begin{itemize}
\item for each link $\frown\ \in \phi_{\gamma_R}$, $\eta \frown \eta$,
  where $\eta $ is a node address in $ \gamma_R$. i.e.
  $\phi_{\gamma_R}$ is a set of reflexive links.
  
\item $\delta_R$ and $\phi_{\gamma_R}$ have some canonical order
  $\prec$.

\item adjunction (similarly substitution) of $\langle \beta_L,
  \beta_R \rangle$ into $\gamma$ is given by \[ \langle
  \gamma'_L, \gamma'_R \rangle = \langle \gamma_L[a_L,
  \beta_L] , \gamma_R[a_R, \beta_R] \rangle\]
  
  and {\em for all} \( \gamma_i \in \delta_\gamma, \beta_i \in
  \phi_{\beta_R} (1 \leq i \leq n) \) (card($\delta_\gamma$) $\geq$ card($\beta_R$))
  \[ \delta_\gamma \sqcup \phi_{\beta_R} \stackrel{def}{=} \frown_{\gamma_1}
  \sqcup \frown_{\beta_1} \cup \ldots \cup \frown_{\gamma_n} \sqcup
  \frown_{\beta_n} \]

where \[ \frown_{\gamma_1} \prec \frown_{\gamma_2}, \ldots, 
         \frown_{\gamma_{n-1}} \prec \frown_{\gamma_n} \]
and
      \[ \frown_{\beta_{R_1}} \prec \frown_{\beta_{R_2}}, \ldots,
         \frown_{\beta_{R_{n-1}}} \prec \frown_{\beta_{R_n}}
      \]

\item $\frown_i \sqcup \frown_j$ is a set of links defined as follows.
  If $a_{L_i} \frown_i a_{R_i}$ and $a_{R_j} \frown_j a_{R_j}$, then
\[ \frown_i \sqcup \frown_j \stackrel{def}{=} \{ a_{L_i} \frown a_{R_i} \} \cup
\{ a_{L_i} \frown a_{R_j} \} \]

\item \( \langle \gamma'_L, \gamma'_R \rangle \) is the new derived
  structure with new set of links $\delta_\gamma \sqcup
  \phi_{\beta_R}$.
\end{itemize}

\end{defn}

$\Phi$ is used to derive synchronized parallelism in $G_R$. The
ordering $\prec$ is simply used to match up the links being shared via
the (non-local) sharing operation $\sqcup$.

This ordering $\prec$ can be defined in terms of node addresses or
``first argument $\prec$ second argument'', i.e. ordering the
arguments of the two predicates being coordinated.

It is important to note that only the links in $\Phi$ are used
non-locally and they are always exhausted in a single adjunction (or
substitution) operation. No links from $\Delta$ are ever inherited
unlike \form{STAG}s. Hence, non-locality is only used in a restricted
fashion for the notion of 'sharing'.

\section{Linguistic Relevance}

To explain how the formalism works consider sentence
\ref{ex:samplederiv}.

\beginsentences
\sitem John cooks and eats beans.
\label{ex:samplederiv}
\endsentences

\begin{figure}[htbp]
  \begin{center}
    \leavevmode
    \psfig{figure=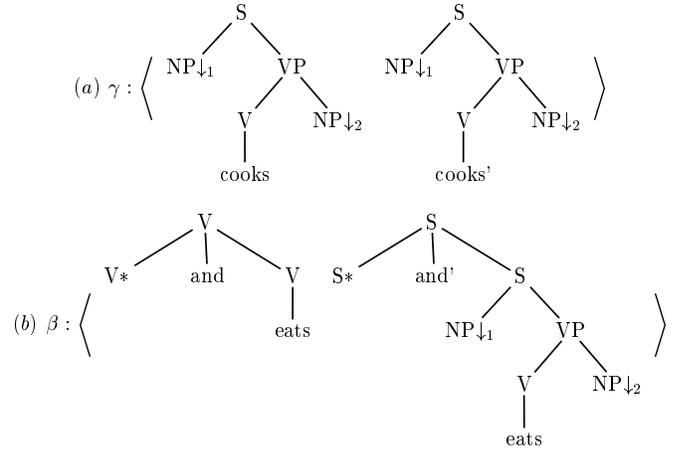,scale=80}
  \end{center}
  \caption{Trees $\gamma$ and $\beta$ from \form{LSTAG} $G$}
  \label{fig:g}
\end{figure}

Consider a \form{LSTAG} $G = \{ \gamma, \beta, \alpha, \upsilon \}$
partially shown in Fig.~\ref{fig:g}(a) and Fig.~\ref{fig:g}(b).
$\alpha$ and $\upsilon$ are analogously defined for {\em John} and
{\em beans} respectively (see Fig.~\ref{fig:tags}).
In Fig.~\ref{fig:g}(a) $\delta_\gamma = \{ 1, 2 \}$%
\footnote{ We are just using numbers $1, 2, \ldots$ to denote the
  links rather than use the Gorn notation to make the trees easier to
  read. Here, link number 1 stands for $1 \frown 1$ and 2 stands for
  $2.2 \frown 2.2$}%
and $\phi_{\gamma_R} = \{ \}$, while for Fig.~\ref{fig:g}(b)
$\delta_\gamma = \{ \}$ and $\phi_{\gamma_R} = \{ 1, 2 \}$. 

It is important to note that our initial motivation about seperating
dependency from the constituency information is highlighted in $\beta$
(see Fig.~\ref{fig:g}(b)) where the first projection will only
contribute information about constituency in a derivation tree while
the second projection will contribute only dependency information in a
derivation tree. We conjecture that this is true for all the
structures defined in an \form{LSTAG}. the kind of questions addressed
in \cite{d-tree95} can perhaps be answered within the framework of
\form{LSTAG}%
\footnote{ In \cite{d-tree95} a new formalism called D-Tree Grammars
  was introduced in order to bring together the notion of derivation
  tree in a \form{TAG} with the notion of dependency
  grammar~\cite{melcuk88}. Perhaps the kind of questions addressed in
  \cite{d-tree95} can also be handled using the current framework.
  Such an application of the formalism would motivate the need for
  trees like $\gamma$ in Fig.~\ref{fig:g} independent of the
  coordination facts since they would be required to get the
  dependencies right.}%
.

\begin{figure*}[htbp]
  \begin{center}
    \leavevmode
    \psfig{figure=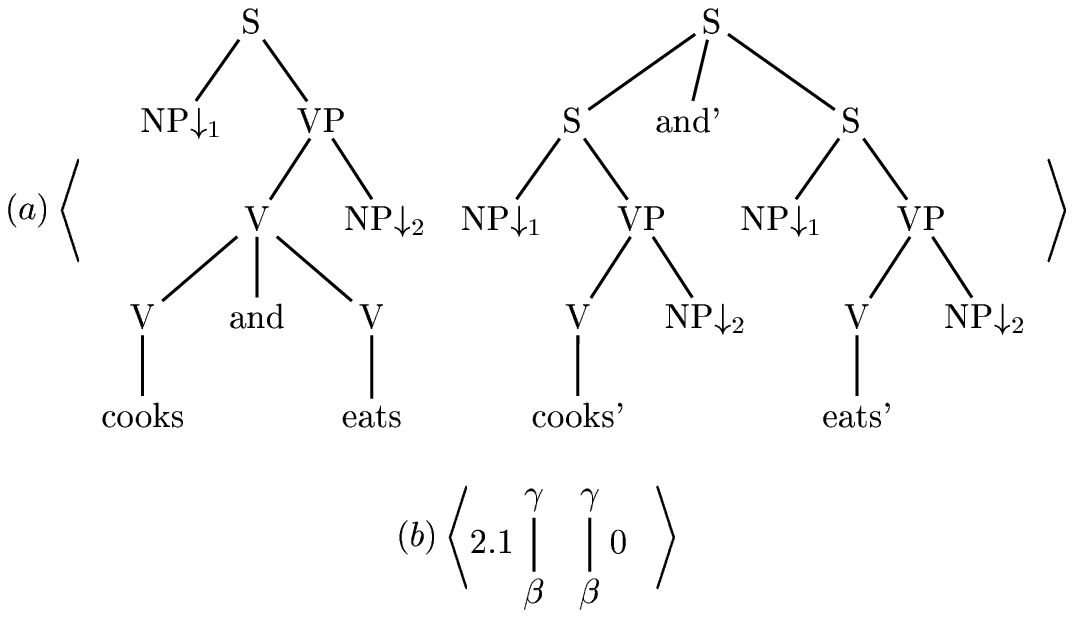,scale=80}
    \label{fig:gb}
    \caption{Derived and derivation structures after $\beta$ adjoins
      into $\gamma$.}
  \end{center}
\end{figure*}

\begin{figure*}[htbp]
  \begin{center}
    \leavevmode
    \psfig{figure=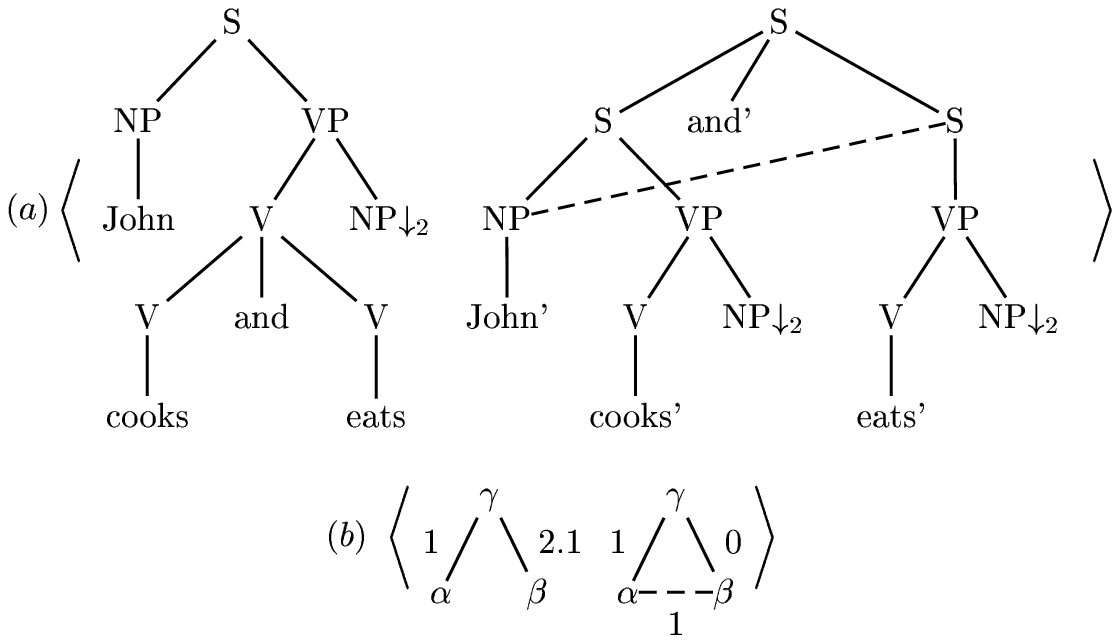,scale=80}
    \caption{Substitution of $\alpha$}
    \label{fig:agb}
  \end{center}
\end{figure*}

The derived structure after $\beta$ adjoins onto $\gamma$ is shown in
Fig.~\ref{fig:gb}(a). Fig.~\ref{fig:agb}(a) shows the derived tree
after the tree $\alpha$ (for {\em John}) substitutes into $\gamma$.
Notice that due to link sharing, substitution is shared, effectively
forming a ``tangled'' derived tree%
\footnote{ While this notion of sharing bears some resemblance to the
  notion of {\em joining node} in the three-dimensional trees used in
  \cite{moltmann92,muadz91} the rules for semantic interpretation of
  the derivations produced in a \form{LSTAG} is considerably less
  obscure than the rules needed to interpret 3D trees; crucially
  because elementary structures in a \form{TAG}-like formalisms are
  taken to be semantically minimal without being semantically void.}%
.  In Figs.~\ref{fig:gb} and~\ref{fig:agb} the derivation trees
are also given (associated with each element). The derivation
structure for the second element in Fig.~\ref{fig:agb}(b) is a
directed acyclic derivation graph which gives us information about
dependency we expect. The derivation tree of the first element in
Fig.~\ref{fig:agb}(b), on the other hand, gives us information about
constituency.

The notion of link sharing is closely related to the schematization of
the coordination rule in \cite{steedman:97} shown below in combinatory
notation.

\begin{eqnarray*}
bxy & \equiv & bxy \\
bfg & \equiv & \lambda x.b(fx)(gx) \\
bfg & \equiv & \lambda x.\lambda y.b(fxy)(gxy) \\
\cdots
\end{eqnarray*}

Link sharing is used to combine the interpretation of the predicate
arguments $f$ and $g$ (e.g. {\em cooks}, {\em eats}) of the
conjunction $b$ with the interpretation of the arguments of those
predicates $x, y, \ldots$.  However, it does this within a
tree-rewriting system, unlike the use of combinators in
\cite{steedman:97}.

\section{Restrictions}

Having defined the formalism of \form{LSTAG}, we now define certain
restrictions on the grammar that can be written in this formalism in
order to capture correctly certain facts about coordinate structures
in English.

\begin{figure*}[htbp]
  \begin{center}
    \leavevmode
    \psfig{figure=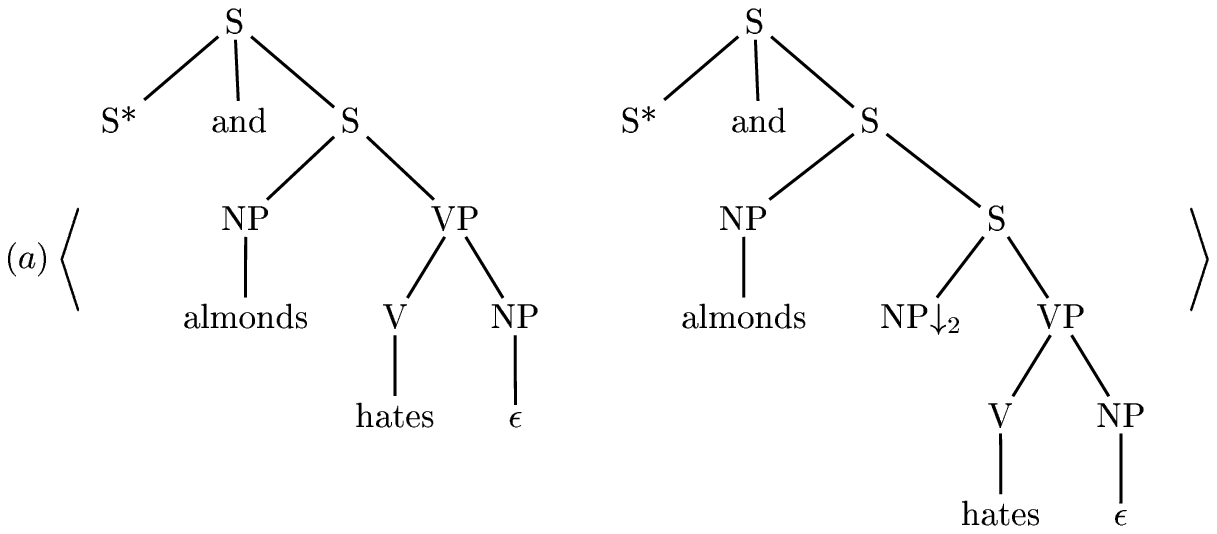,scale=80}
    \caption{Discontiguous elementary structure}
    \label{fig:noncont}
  \end{center}
\end{figure*}

For instance, we need to prohibit elementary structures like the one
in Fig.~\ref{fig:noncont} because they give rise to ungrammatical
sentences like \ref{ex:ungram}.

\beginsentences
\sitem *Peanuts John likes and almonds hates. \cite{joshi90}
\label{ex:ungram}
\endsentences

However, such restrictions in the context of \form{TAG}s have been
discussed before. \cite{joshi90} rules out \ref{ex:ungram} by stating
a requirement on the lexical string spelled out by the elementary
tree. If the lexical string spelled out is not contiguous then it
cannot coordinate. This requirement is stated to be a phonological
condition and relates the notion of an {\em intonational phrase} (IP)
to the notion of appropriate fragments for coordination (in the spirit
of \cite{steedman:ms}). It is important to note that the notions of
phrase structure for coordination and intonational phrases defined in
\cite{joshi90} for \form{TAG} are not identical, whereas they {\em
  are} identical for CCG~\cite{steedman:ms}.

We can state an analogous restriction on the formation of elementary
structures in a \form{LSTAG}, one that is motivated by the notion of
link sharing. The left element of an elementary structure in a
\form{LSTAG} cannot be composed of discontinuous parts of the right
element. For example, in Fig.~\ref{fig:noncont} the segment $[_S
[_{NP_\downarrow} ] [_{VP} ] ]$ from the right element has been excised
in the left element. This restriction corresponds to the notion that
the left element of a structure in a \form{LSTAG} represents
constituency.

\section{Conclusion}

We have presented a new tree-rewriting formalism called Link-Sharing
Tree Adjoining Grammar (\form{LSTAG}) which is a variant of
synchronous TAGs (\form{STAG}). Using \form{LSTAG} we defined an
approach towards coordination where linguistic dependency is
distinguished from the notion of constituency. Appropriate
restrictions on the nature of elementary structures in a \form{LSTAG}
were also defined. Such an approach towards coordination that
explicitly distinguishes dependencies from constituency gives a better
formal understanding of its representation when compared to previous
approaches that use tree-rewriting systems which conflate the two
issues (see \cite{Joshi91,sarkarjoshi96}). The previous approaches had
to represent coordinate structures either with unrooted trees or by
performing structure merging on the parse tree. Moreover, the
linguistic analyses presented in \cite{Joshi91,sarkarjoshi96} can be
easily adopted in the current formalism.

{\small
\bibliographystyle{fullname}
\bibliography{mol5}}

\begin{thebibliography}{}

\bibitem[\protect\citename{Abeill\'e}1992]{abeille92}
Abeill\'e, Anne.
\newblock 1992.
\newblock {Synchronous TAGs and French Pronominal Clitics}.
\newblock In {\em {Proc. of COLING-92}}, pages 60--66, Nantes, Aug 23--28.

\bibitem[\protect\citename{Abeill\'e}1994]{abeille94}
Abeill\'e, Anne.
\newblock 1994.
\newblock {Syntax or Semantics? Handling Nonlocal Dependencies with MCTAGs or
  Synchronous TAGs}.
\newblock {\em Computational Intelligence}, 10(4):471--485.

\bibitem[\protect\citename{Engelfriet, Rozenberg, and
  Slutzki}1980]{engelfriet:etal80}
Engelfriet, J., G.~Rozenberg, and G.~Slutzki.
\newblock 1980.
\newblock Tree transducers, {$L$} systems, and two-way machines.
\newblock {\em Journal of Computer and System Science}, 43:328--360.

\bibitem[\protect\citename{Frank}1992]{frank92}
Frank, Robert.
\newblock 1992.
\newblock {\em {Syntactic locality and Tree Adjoining Grammar: grammatical,
  acquisition and processing perspectives}}.
\newblock {Ph.D.} thesis, University of Pennsylvania,IRCS-92-47.

\bibitem[\protect\citename{Joshi and Schabes}1991]{joshi-schabes91}
Joshi, A. and Y.~Schabes.
\newblock 1991.
\newblock Tree adjoining grammars and lexicalized grammars.
\newblock In M.~Nivat and A.~Podelski, editors, {\em Tree automata and
  languages}. North-Holland.

\bibitem[\protect\citename{Joshi}1990]{joshi90}
Joshi, Aravind.
\newblock 1990.
\newblock {Phrase Structure and Intonational Phrases: Comments on the papers by
  Marcus and Steedman}.
\newblock In G.~Altmann, editor, {\em Computational and Cognitive Models of
  Speech}. MIT Press.

\bibitem[\protect\citename{Joshi and Schabes}1991]{Joshi91}
Joshi, Aravind and Yves Schabes.
\newblock 1991.
\newblock Fixed and flexible phrase structure: {Coordination in Tree Adjoining
  Grammar}.
\newblock In {\em Presented at the {DARPA} Workshop on Spoken Language
  Systems}, Asilomar, CA.

\bibitem[\protect\citename{Joshi, Levy, and Takahashi}1975]{joshi75}
Joshi, Aravind~K., L.~Levy, and M.~Takahashi.
\newblock 1975.
\newblock {Tree Adjunct Grammars}.
\newblock {\em Journal of Computer and System Sciences}.

\bibitem[\protect\citename{Kroch}1987]{kroch87}
Kroch, A.
\newblock 1987.
\newblock Subjacency in a tree adjoining grammar.
\newblock In A.~Manaster-Ramer, editor, {\em Mathematics of Language}. J.
  Benjamins Pub. Co., pages 143--172.

\bibitem[\protect\citename{Mel'\^{c}uk}1988]{melcuk88}
Mel'\^{c}uk, I.
\newblock 1988.
\newblock {\em {Dependency Syntax: Theory and Practice}}.
\newblock State University of New York Press, Albany.

\bibitem[\protect\citename{Moltmann}1992]{moltmann92}
Moltmann, Friederike.
\newblock 1992.
\newblock {On the Interpretation of Three-Dimensonal Syntactic Trees}.
\newblock In Chris Barker and David Dowty, editors, {\em {Proc. of SALT-2}},
  pages 261--281, May 1-3.

\bibitem[\protect\citename{Muadz}1991]{muadz91}
Muadz, H.
\newblock 1991.
\newblock {\em {A Planar Theory of Coordination}}.
\newblock {Ph.D.} thesis, University of Arizona, Tucson, Arizona.

\bibitem[\protect\citename{Rambow and Joshi}1992]{rambowjoshi92}
Rambow, O. and A.~Joshi.
\newblock 1992.
\newblock A formal look at dependency grammars and phrase-structure grammars,
  with special consideration to word-order phenomena.
\newblock In {\em Intern. Workshop on the Meaning-Text Theory}, pages 47--66,
  Arbeitspapiere der GMD 671. Darmstadt.

\bibitem[\protect\citename{Rambow and Satta}to appear]{sattarambow:toappear}
Rambow, O. and G.~Satta.
\newblock to appear.
\newblock Independent parallelism in finite copying parallel rewriting systems.
\newblock {\em Theor. Comput. Sc.}

\bibitem[\protect\citename{Rambow, Vijay-Shanker, and Weir}1995]{d-tree95}
Rambow, O., K.~Vijay-Shanker, and D.~Weir.
\newblock 1995.
\newblock {D-Tree Grammars}.
\newblock In {\em Proceedings of the 33rd Meeting of the ACL}.

\bibitem[\protect\citename{Sarkar and Joshi}1996]{sarkarjoshi96}
Sarkar, Anoop and Aravind Joshi.
\newblock 1996.
\newblock Coordination in {TAG}: Formalization and implementation.
\newblock In {\em Proceedings of the 16th International Conference on
  Computational Linguistics (COLING'96)}, Copenhagen.

\bibitem[\protect\citename{Shieber}1994]{shieber94}
Shieber, S.
\newblock 1994.
\newblock Restricting the weak generative capacity of synchronous tree
  adjoining grammars.
\newblock {\em Computational Intelligence}, 10(4):371--385, November.

\bibitem[\protect\citename{Shieber and Schabes}1990]{shieber90}
Shieber, Stuart and Yves Schabes.
\newblock 1990.
\newblock {Synchronous Tree Adjoining Grammars}.
\newblock In {\em {Proceedings of the $13^{th}$ International Conference on
  Computational Linguistics (COLING'90)}}, {Helsinki, Finland}.

\bibitem[\protect\citename{Steedman}1985]{steedman85}
Steedman, Mark.
\newblock 1985.
\newblock Dependency and coordination in the grammar of {Dutch} and {English}.
\newblock {\em Language}, 61:523--568.

\bibitem[\protect\citename{Steedman}1997a]{steedman:ms}
Steedman, Mark.
\newblock 1997a.
\newblock {Information Structure and the Syntax-Phonology Interface}.
\newblock manuscript. Univ. of Pennsylvania.

\bibitem[\protect\citename{Steedman}1997b]{steedman:97}
Steedman, Mark.
\newblock 1997b.
\newblock {\em Surface {S}tructure and {I}nterpretation: {U}nbounded and
  {B}ounded {D}ependency in {C}ombinatory {G}rammar}.
\newblock Linguistic Inquiry monograph. MIT Press.

\end{thebibliography}

\end{document}